\documentclass[11pt,a4paper]{article}
\usepackage[nohyperref]{acl2019}
\usepackage{times}
\usepackage{latexsym}

\usepackage{url}

\aclfinalcopy 

\usepackage{url}
\usepackage{bbm}
\usepackage{makecell}
\usepackage{pifont}
\usepackage{multirow}
\usepackage{booktabs}
\usepackage{tabularx}
\usepackage{caption}
\usepackage{comment}
\usepackage[title]{appendix}
\usepackage{dblfloatfix}
\usepackage{adjustbox}
\usepackage[warn]{textcomp}
\usepackage{endnotes}

\usepackage{graphicx}
\graphicspath{ {./images/} }

\usepackage{float}
\restylefloat{table}

\usepackage{subfigure}
\usepackage{boxedminipage}

\newcommand{\xmark}{\ding{55}}

\newcolumntype{Y}{>{\centering\arraybackslash}X} 

\usepackage{amsmath,amsfonts,bm}

\def\eqref#1{equation~\ref{#1}}

\def\1{\bm{1}}

\DeclareMathAlphabet{\mathsfit}{\encodingdefault}{\sfdefault}{m}{sl}
\SetMathAlphabet{\mathsfit}{bold}{\encodingdefault}{\sfdefault}{bx}{n}

\newcommand{\sigmoid}{\sigma}

\title{Keeping Notes: Conditional Natural Language Generation\\ with a Scratchpad Mechanism}
\author{
  Ryan Y. Benmalek$^{\dagger*}$,
  Madian Khabsa$^{\ddagger*}$,
  Suma Desu$^{\mathparagraph*}$,
  Claire Cardie$^{\dagger}$,
  Michele Banko$^{\mathsection*}$
  \\
  $^{\dagger}$Cornell University,
  $^{\ddagger}$Facebook,
  $^{\mathparagraph}$Independent ,
  $^{\mathsection}$Sentropy Technologies \\
  \texttt{ryanai3@cs.cornell.edu, mkhabsa@fb.com,} \\ \texttt{desuma24@gmail.com, mbanko@sentropy.io, cardie@cs.cornell.edu}
}
\date{}

\begin{document}
\maketitle
\makeatletter
\def\blfootnote{\gdef\@thefnmark{}\@footnotetext}
\makeatother
\blfootnote{$^{*}$ Work performed while at Apple.}

\begin{abstract}
We introduce the \textit{Scratchpad Mechanism}, a novel addition to the sequence-to-sequence (seq2seq) neural network architecture and demonstrate its effectiveness in improving the overall fluency of seq2seq models for natural language generation tasks. By enabling the decoder at each time step to write to all of the encoder output layers, \textit{Scratchpad} can employ the encoder as a ``scratchpad" memory to keep track of what has been generated so far and thereby guide future generation.   We evaluate \textit{Scratchpad} in the context of three well-studied natural language generation tasks --- Machine Translation, Question Generation, and Text Summarization --- and obtain state-of-the-art or comparable performance on standard datasets for each task. Qualitative assessments in the form of human judgements (question generation), attention visualization (MT), and sample output (summarization) provide further evidence of the ability of \textit{Scratchpad} to generate fluent and expressive output.  
\end{abstract}

\section{Introduction}
The sequence-to-sequence neural network framework (seq2seq) \cite{Seq2Seq} has been successful in a wide range of tasks
in natural language processing, from machine translation \cite{Attention} and semantic parsing \cite{dong2016language} to summarization \cite{Nallapati2016AbstractiveTS, see2017get}. Despite this success, seq2seq models are known to often exhibit an overall lack of fluency in the natural language output produced: problems include lexical repetition, under-generation in the form of partial phrases and lack of specificity (often caused by the gap between the input and output vocabularies) \cite{xie2017neural}. Recently, a number of \emph{task-specific attention} variants have been proposed to deal with these issues: \newcite{see2017get} introduced a coverage mechanism \cite{Coverage} to deal with repetition and over-copying in summarization, \newcite{Hua2018NeuralAG} introduced a method of attending over keyphrases to improve argument generation, and \newcite{Kiddon2016GloballyCT} introduced a method that attends to an agenda of items to improve recipe generation.
Perhaps not surprisingly, general-purpose attention mechanisms  targeting individual problems from the list above have also begun to be developed. Copynet \cite{Copynet} and pointer-generator networks \cite{Vinyals2015PointerN}, for example, aim to reduce input-output vocabulary mismatch and, thereby, improve specificity, while the coverage-based techniques of \newcite{Coverage} tackle repetition and under-generation.  These techniques,  however, often require significant hyperparameter tuning and are purposely limited to fixing a specific problem in the generated text. 

We present here a general-purpose addition to the standard seq2seq framework that aims to simultaneously tackle all of the above issues.  In particular,  we propose \textit{Scratchpad}, a novel write mechanism that allows the decoder to ‘keep notes’ on its past actions (i.e., generation, attention, copying) by directly modifying encoder states.
The \textit{Scratchpad} mechanism essentially lets the decoder 
 more easily keep track of what the model has focused on and copied from the input in the recent past, as well as what it has produced thus far as output. Thus, \textit{Scratchpad}  can alternatively be viewed as an external memory initialized by the input, or as an input re-encoding step that takes into account past attention and generation.
%

To demonstrate general(izable) improvements on conditional natural language generation problems broadly construed, 
we instantiate \textit{Scratchpad} for three well-studied generation tasks --- Machine Translation, Question Generation, and Summarization --- and evaluate it on a diverse set of datasets.  These tasks exhibit a variety of input modalities (structured and unstructured language) and typically have required a variety of computational strategies to perform well (attention, pointing, copying).
We find, for each task, that \textit{Scratchpad} attains 
improvements over several strong baselines: Sequence-to-Sequence with attention \cite{Seq2Seq,Attention}, copy-enhanced approaches \cite{Copynet,Vinyals2015PointerN}, and coverage-enhanced approaches \cite{Coverage,see2017get}. \textit{Scratchpad} furthermore obtains state-of-the-art performance for each task.
Qualitative assessments in the form of human judgements (question generation), attention visualization (MT) and sample output (summarization) provide further evidence of the ability of \textit{Scratchpad} to generate fluent and expressive output.



\section{Background}
\textit{Scratchpad} builds upon a standard attention-based seq2seq neural architecture \cite{Attention} comprised of (a) an \textit{encoder} that operates token-by-token over the \textit{input}, (b) a \textit{decoder} that produces the \textit{output}, and (c) an \textit{attention} mechanism that allows the decoder to focus on different parts of the input.  In the subsections below, we first briefly review this architecture (we assume the reader is familiar with the framework). In Section~\ref{scratchpad}, we introduce the \textit{Scratchpad} mechanism.

\newcommand{\mbf}[1]{\mathbf{#1}}
\newcommand{\h}{\mbf{h}}
\newcommand{\s}{\mbf{s}}
\newcommand{\at}{\mbf{a}}
\newcommand{\yhat}{\hat{y}}
\newcommand{\ut}{\mbf{u}}

\newcommand{\enchs}{[\h_1,...,\h_n]}
\newcommand{\attn}{\text{attn}}
\newcommand{\scor}{\text{score}}
\newcommand{\aread}{\mbf{c}^i}
\newcommand{\softmx}{\text{softmax}}
\newcommand{\RNN}{\text{RNN}}
\newcommand{\MLP}{\text{MLP}}
\newcommand{\htan}{\text{tanh}}

\paragraph{Encoder}
Let $X = [x_1,...,x_n]$ denote an input sequence of length $N$ where $x_i$ is the $i$-th token. The encoder is a recurrent neural network (RNN) that produces in its final layer a sequence of hidden states $\enchs = \text{RNN}(\{\mbf{x_1},...,\mbf{x_n}\})$.  These can be viewed as a sequence of token-level feature vectors learned from the input.

\paragraph{Decoder}
Let the decoding sequence be indexed by the superscript $i$. The decoder is an RNN whose initial hidden state $\s^0$ is set to the final state(s) of the of the encoder.

\paragraph{Attention} 
At every decoding step $i$, an attention mechanism, i.e., an attentive read (often termed \textit{attentional context}) ($\aread$), is derived from the encoder output states ($\enchs$). 
Concretely, attention is applied by first computing a {\it score} for each encoder output, $\mbf{h}_t$: 
\begin{equation}
\scor^i_t =  \mbf{W_1}(\mbf{W_2}[\s^i,\h_t]^\mbf{T})
\end{equation}
where weight matrices $\mbf{W_1}$ and $\mbf{W_2}$ are learned parameters.
These scores, $\scor^i_{1..T}$, are then normalized into a probability distribution:
\begin{equation}
\at^i = \softmx(\scor^i_{1..T})
\end{equation}
The \textit{attentive read} operation is then the weighted average of encoder outputs according to this distribution, which allows the decoder to focus on different parts of the input at different timesteps $i$:
\begin{equation}\label{eq:attn_read}
\aread = \sum_{t=1}^T(\at^i_t * \h_t)
\end{equation}

\section{Scratchpad Mechanism}
\label{scratchpad}
The above attention mechanism has been widely successful in many generation tasks but the quality of generated text still suffers from caveats and requires significant tuning. We augment attention with a \textit{Scratchpad} mechanism to introduce higher quality generated text with less effort.
Intuitively, \textit{Scratchpad} adds one simple step to the decoder: treating the encoder output states,  $\enchs$, as a \textit{scratchpad}, thus it writes to them as if the set of states were an external memory. Exactly how this is done is described next.

Without \textit{Scratchpad}, the decoder's workflow at every output timestep step $\mathrm{i}$ is as follows:

\noindent
1. \textit{\textbf{Read}} attentively ($\aread$) from the encoder outputs ($\enchs$) using the current state, $\s^i$. 

\noindent
2. \textit{\textbf{Update}} $\s^i$ using the most recently generated output token, $y^{i-1}$,  and the results of the attentive read ($\aread$). 

\noindent
3.
\textit{\textbf{Output}} a distribution over the output vocabulary $\mbf{\yhat}^i$.

\noindent
\textit{Scratchpad} simply adds a fourth step:

\noindent
4. \textit{\textbf{Write}} an update ($\ut^i$) to the encoder outputs ($\enchs$) in an attentive fashion ($\alpha^i_{1..T}$), treating the encoder outputs as if they were cells in an external memory.

\begin{figure}[tbhp]
\centering
\includegraphics[height=3in]{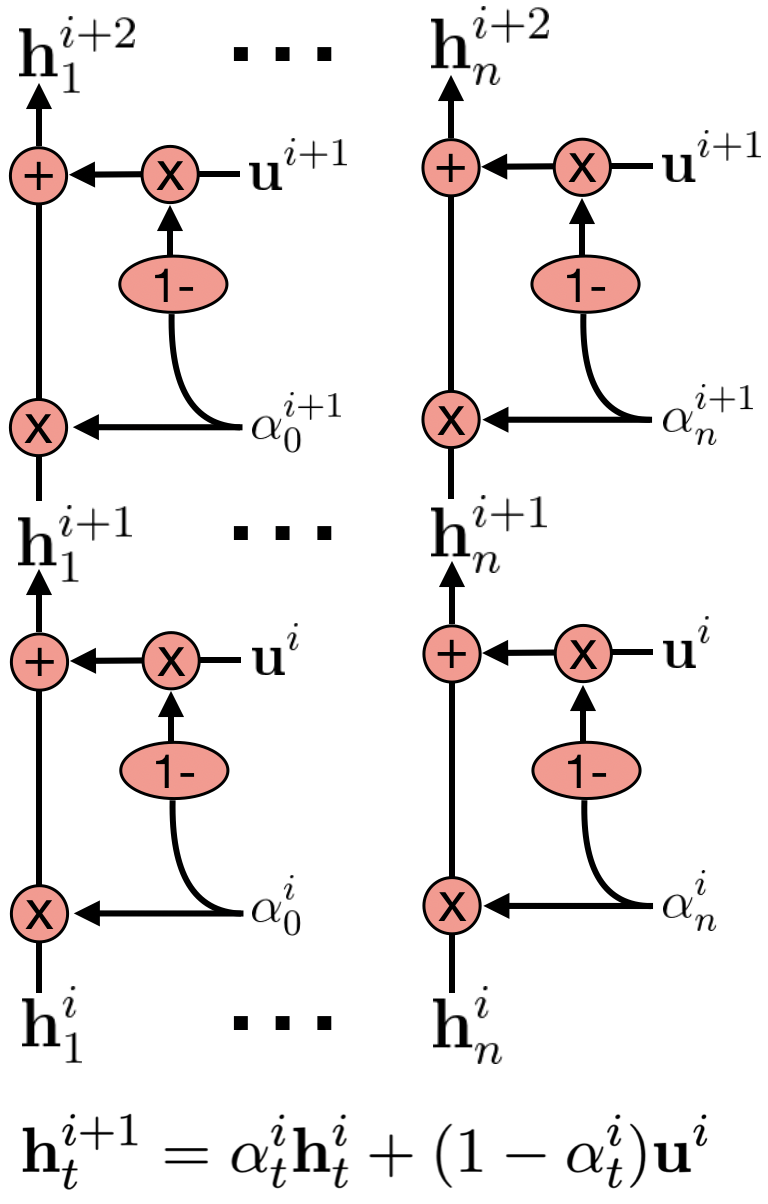}
\caption{The \textit{Scratchpad Mechanism} first computes the update probability ($\alpha^i_t$) for each encoder state according to Eq.~\ref{eq:write_attn}, then computes a global update $\ut^i$ according to Eq.~\ref{eq:calc_update}, and finally updates the encoder states according to Eq.~\ref{eq:scratchpad_write}.}
\label{fig:model_diagram}
\end{figure}
More specifically, to calculate both the write-attention and the update, \textit{Scratchpad} uses the concatenation of the decoder state after steps 1-3 ($\s^{i+1}$) and the attentive read ($\aread$): 
\begin{equation}\label{eq:scratchpad_write}
\h^{i+1}_t = \alpha^i_t \h^i_t + (1-\alpha^i_t) \ut^i
\end{equation}
\begin{equation}\label{eq:write_attn}
\alpha^i_t = \sigmoid(f_{\alpha}([\s^{i+1}, \aread, \h^i_t]))
\end{equation}
\begin{equation}\label{eq:calc_update}
\ut^i = \htan(f_u([\s^{i+1};\aread]))
\end{equation}
In essence, the \textit{Scratchpad} consists of two components. The first determines what 'notes' to keep ($\ut^i$). The second is similar to the 'forget' gate in an LSTM, where the network decides how much to overwrite a cell ($1-\alpha^i_t$) versus how much to keep past information ($\alpha^i_t$) for that cell. These two components are used in concert (see Fig.~\ref{fig:model_diagram}) to provide new encoder states ($h^{i+1}_t$) to the decoder at each decoding timestep ($i$).
Tanh is used to ensure that $\h^{i+1}_t$ remains in the range $[-1, 1]$, since $\h^{i}_t \in [-1, 1]$ as $\enchs^0$ is the hidden states of a GRU or LSTM. 
We parametrize $f_{\alpha}$ and $f_u$ as an MLP.
Figure \ref{fig:model_diagram} shows the outline of the scratchpad mechanism update at multiple timesteps.

\section{Experiments}
In this section we describe experimental setup and results for  Machine Translation, Question Generation, and Summarization tasks which exhibit a variety of input modalities and strategies required to perform well. We work with structured and unstructured language and several sequence to sequences  architectures i.e. attention, pointing, and copying. 
Machine translation is a canonical sequence-to-sequence task where pairwise word-level or phrase-level generation is expected. Question Generation from logical forms requires reasoning about the syntax, parse tree, and vocabulary of the input sequence to infer the meaning of the logical form program and utilize copy-mechanism to 
copy entities. Lastly, summarization requires understanding both the global and the local context of a sentence 
within a document, identifying spans that are informative and diverse, and generating coherent representative summaries. Demonstrating a single mechanism that reaches state of the art on such a diverse set of natural language tasks underscores the generalizability of our technique, particularly given the large range in number of training examples (3k, 56k, 153k, 287k) across datasets.

\subsection{Translation} \label{translation}
We evaluate on the IWLST 14 English to German and Spanish to English translation datasets \cite{IWSLT} as well as the IWSLT 15 \cite{IWSLT} English to Vietnamese translation dataset.
For IWSLT14 \cite{IWSLT}, we compare to the models evaluated by \newcite{He2018LayerWiseCB}, which includes a transformer \cite{Vaswani2017AttentionIA} and RNN-based models \cite{Attention}. For IWSLT15, we primarily compare to GNMT \cite{Wu2016GooglesNM}, which incorporates Coverage \cite{Coverage}.
Table \ref{table:translation} shows BLEU scores of 
our approach on 3 IWSLT translation tasks along with 
reported results from previous work. Our approach 
achieves state-of-the-art or comparable results on all 
datasets.

\begin{table}[H]
\centering
\begin{adjustbox}{width={\columnwidth}}
\begin{tabular}{l c c c}
\toprule
Model & \multicolumn{2}{c}{IWSLT14} & IWSLT15 \\
\cmidrule(lr){2-3} \cmidrule(lr){4-4} 
 & De\textrightarrow En & Es \textrightarrow En & En\textrightarrow Vi \\


\midrule
MIXER & $21.83$ & \xmark & \xmark\\
AC + LL & $28.53$ & \xmark & \xmark \\
NPMT & $29.96$ & \xmark & 28.07 \\
\midrule
Stanford NMT & \xmark & \xmark & $26.1$\\
Transformer (6 layer)& $32.86$ & $38.57$ & \xmark \\ 
Layer-Coord (14 layer) & $35.07$ & $40.50$ & \xmark \\ 
Scratchpad (3 layer) &  $\mathbf{35.08}$ & $\mathbf{40.92}$ & $\mathbf{29.59}^*$\\
\bottomrule
\end{tabular}
\caption{Performance for non-scratchpad models are taken from \protect\newcite{He2018LayerWiseCB} except Stanford NMT \protect\cite{Luong2015NeuralMT}. $*$: model is 2 layers.}
\label{table:translation}
\end{adjustbox}
\end{table}
\begin{table*}[t]
\centering
\begin{tabular}{l l c c c c c c}

& \multirow{2}[3]{*}{Model} & \multicolumn{3}{c}{Per-Sentence} & \multicolumn{3}{c}{Corpus-Level} \\

\cmidrule(lr){3-5}
\cmidrule(lr){6-8}

& & Bleu & Meteor & Rouge-L & Bleu & Meteor & Rouge-L \\
\cmidrule[1pt](lr){2-8}
\multirow{4}{*}{\rotatebox{90}{WebQSP}} & Baseline & $7.51$ & $23.9$ & $47.1$ & $17.96$ & $22.9$ & $47.13$ \\

\cmidrule(lr){2-8}

& Copynet & $6.89$ & $27.1$ & $52.5$ & $17.42$ & $26.03$ & $52.56$ \\
& Copy + Coverage & $14.55$ & $33.7$ & $58.9$ & $26.78$ & $30.86$ & $58.91$ \\
& Copy + Scratchpad & $\mathbf{15.29}$ & $\mathbf{34.7}$ & $\mathbf{59.5}$ & $\mathbf{27.64}$ & $\mathbf{31.49}$ & $\mathbf{59.44}$ \\

\cmidrule[1pt](lr){2-8}
\multirow{4}{*}{\rotatebox{90}{WikiSQL}} & Baseline & $9.94$ & $26.71$ & $47.96$ & $17.34$ & $25.34$ & $47.96$ \\

\cmidrule(lr){2-8}

& Copynet & $8.04$ & $24.66$ & $46.82$ & $15.11$ & $23.53$ & $46.82$ \\
& Copy + Coverage & $15.76$ & $34.04$ & $54.94$ & $25.01$ & $32.38$ & $54.94$ \\
& Copy + Scratchpad & $\mathbf{16.89}$ & $\mathbf{34.47}$ & $\mathbf{55.69}$ & $\mathbf{26.10}$ & $\mathbf{32.76}$ & $\mathbf{55.69}$ \\

\cmidrule[1pt](lr){2-8}

\end{tabular}
\caption{Methods allowing the model to keep track of past attention (\textit{Coverage}, \textit{Scratchpad}) significantly improve performance when combined with a copy mechanism. The \textit{Scratchpad Encoder} achieves the best performance.}
\label{table:quant_metrics}
\end{table*}
\paragraph{Experimental Details}
For IWSLT14, our encoder is a 3-layer Bi-LSTM \cite{LSTM}, where outputs are combined by concatenation, and the decoder is a 3-layer LSTM as well. For IWSLT15 the encoder and decoder are 2-layers. We follow \newcite{Luong2015EffectiveAT}, using the 'general' score function, input feeding, and combining the attentional context and hidden state. Since we use input feeding, Steps (1) and (2) in Section~\ref{scratchpad} are switched. All our models have a hidden size of $512$ (for the LSTM and any MLP's). The internal layers of the decoder are residual, adding their output to their input and putting it through Layer Normalization \cite{Ba2016LayerN}. Sentences were encoded using byte-pair encoding \cite{bpe}, with a shared source-target vocabulary of $10,000$ for De\textrightarrow En and Es\textrightarrow En (En \textrightarrow Vi uses words as tokens to be comparable to \newcite{Wu2016GooglesNM}). Source and target word embeddings are dimension $128$. We use dropout \cite{dropout} in the encoder and decoder with a probability of $0.1$. We use the Adam optimizer \cite{Adam}, with an initial learning rate of $0.002$.We train for 30/20 epochs for IWSLT14/15, decaying the learning rate by a factor of $0.7$ whenever the validation loss does not improve from the last epoch. Each training batch contained at most 2000 source or target tokens. We use label smoothing with $\epsilon_{ls} = 0.1$ \cite{Szegedy2016RethinkingTI}. We average the last 5 epochs to obtain the final model and run with a beam of size 4.

\subsection{Question Generation}
We use the task of \textit{question generation}: Given a \textit{structured representation} of a query against a knowledge base or a database (e.g. a \textit{logical form}), produce the corresponding natural language question.
We use two datasets consisting of \textit{(question, logical form)} pairs: WebQuestionsSP \cite{WebQSP} (a standard dataset for semantic parsing, where the logical form is in SPARQL), and WikiSQL \cite{zhong2017seq2sql} (where the logical form is SQL). Both datasets are small, with the former having 3098 training and 1639 testing examples, and the latter being an order of magnitude larger with 56346 training and 15873 testing examples.

We evaluate metrics at both a \emph{corpus} level (to indicate how \emph{natural} output questions are) and at a \emph{per-sentence} level (to demonstrate how well output questions exactly \emph{match} the gold question). BLEU \cite{Bleu}, ROUGE \cite{Rouge} are chosen for precision and recall-based metrics. METEOR \cite{Meteor} is chosen to deal with stemming and synonyms.

We noticed that many tokens that appear in the logical form are also present in the natural language form for each example. In fact, nearly half of the tokens in the question appear in the corresponding SPARQL of the WebQuestionSP dataset \cite{WebQSP}, implying that a network with the ability to copy from the input could see significant gains on the task. Accordingly, we compare our \textit{Scratchpad Mechanism} against three baselines: (1) Seq2Seq, (2) \textit{Copynet} and (3) \textit{Coverage}, a method introduced by \newcite{Coverage} that aims to solve attention-related problems. 
Seq2Seq is the standard approach introduced in \newcite{Seq2Seq}. 
The \textit{Copynet} \cite{CoreQA} baseline additionally gives the Seq2Seq model the ability to copy vocabulary from the source to the target.

From Table \ref{table:quant_metrics} it is clear that our approach, \textit{Scratchpad} outperforms all baselines on all the metrics.

\begin{table*}[t!]
\centering
\begin{tabular}{l c c c c c}
\toprule

\multirow{2}[3]{*}{Model} & \multicolumn{3}{c}{Rouge} & \multicolumn{2}{c}{Meteor} \\

\cmidrule(lr){2-4}
\cmidrule(lr){5-6}

& 1 & 2 & L & exact match & +stem/syn/para \\
\midrule
Pointer-generator & $36.44$ & $15.66$ & $33.42$ & $15.35$ & $16.65$\\
\newcite{see2017get} & $39.53$ & $17.28$ & $36.38$ & $\mathbf{17.32}$ & $\mathbf{18.72}$\\
Scratchpad & $\mathbf{39.65}$ & $\mathbf{17.61}$ & $\mathbf{36.62}$ & $17.26$ & $18.63$\\
\midrule
CopyTransformer + Coverage Penalty & $39.25$ & $17.54$ & $36.45$ & \xmark & \xmark \\
Pointer-Generator + Mask Only & $37.70$ & $15.63$ & $35.49$ & \xmark & \xmark \\
\midrule
Bottom-up \cite{Gehrmann2018BottomUpAS} & $\mbf{41.22}$ & $\mbf{18.68}$ & $\mbf{38.34}$ & \xmark & \xmark \\
\bottomrule
\end{tabular}
\caption{The middle third of the table contains the end-to-end models performing the best from \protect\cite{Gehrmann2018BottomUpAS}, while the bottom section contains the current state-of-the-art which involves a 2-step training process and is not end-to-end. Scratchpad establishes a state of the art for end-to-end models on summarization without Reinforcement Learning on ROUGE, while remaining competitive with \protect\newcite{see2017get} on METEOR. Additionally, Scratchpad does not use an auxiliary loss as in \protect\newcite{see2017get} or the middle third of the table. \protect\newcite{Gehrmann2018BottomUpAS} do not evaluate on METEOR.}
\label{table:summarization}
\end{table*}


\paragraph{Experimental Details}
Our encoder is a 2-layer bi-directional GRU where outputs are combined by concatenation, and our decoder is a 2-layer GRU. We use the attention mechanism from \ref{translation}. We train all models for $75$ epochs with a batch size of $32$, a hidden size of $512$ (for the GRU and any MLP's), and a word vector size of $300$. Dropout is used on every layer except the output layer, with a drop probability of $0.5$. Where Glove vectors \cite{Glove} are used to initialize word vectors, we use 300-dimensional vectors trained on Wikipedia and Gigaword ($6B.300D$). We use the Adam optimizer with a learning rate of $1e^{-4}$ and we do teacher forcing \cite{TeacherForcing} with probability $0.5$. These hyperparameters were tuned for our Seq2Seq baselines and held constant for the rest of the models. The vocabulary consists of all tokens appearing at least once in the training set.

\subsection{Summarization}
We use the CNN/Daily Mail dataset \cite{Hermann2015TeachingMT,Nallapati2016AbstractiveTS} as in \newcite{see2017get}. The dataset consists of 287,226 training, 13,368 validation, and 11,490 test examples. Each example is an online news article (781 tokens on average) along with a multi-sentence summary (56 tokens, 3.75 sentences on average). As in \newcite{see2017get} we operate on the original non-anonymized version of the data.

We follow \newcite{see2017get} in evaluating with ROUGE \cite{Rouge} and METEOR \cite{Meteor}. We report $\mathrm{F_1}$ scores for ROUGE-1, ROUGE-2, and ROUGE-LCS (measuring word, bigram, and longest-common-subsequence overlap, respectively) and  we report METEOR in exact and full mode.
We compare to the pointer-generator baseline and the coverage variant introduced by \newcite{see2017get}. \newcite{see2017get} use a multi-step training procedure for the coverage component to improve performance where a pointer-generator model is first trained without the coverage component for a large number of iterations, then trained with the component and a tuned auxiliary coverage loss, finding that the auxiliary loss and pre-training the network without coverage are required to improve performance. As demonstrated in Tab.~\ref{table:summarization}, with \textit{Scratchpad}, we are able to improve performance over \newcite{see2017get} in all the Rouge metrics, statistically significant for Rouge-2 and Rouge-L, while remaining comparable in METEOR. We reach this performance with half of the training iterations, no pretraining, and without the additional memory outlay and model complexity of including an auxiliary loss.

\paragraph{Experimental Details}
We use the same setup as in \newcite{see2017get}: The encoder is a single-layer bi-directional LSTM where outputs are combined by concatenation, and the decoder consists of a single-layer LSTM. The encoder states modified by the \textit{scratchpad mechanism} are the outputs of the LSTM at every timestep, i.e. the 'hidden' state. We use the same attention mechanism as in \newcite{see2017get} to calculate the attentive read and the attentive write probabilities $\alpha^i_t$ for the \textit{scratchpad mechanism}.
\newcite{see2017get} introduce a multi-step training procedure where a pointer-generator model is first trained with the vanilla cross-entropy objective for 230k iterations. Then the coverage component is added and the full model is further trained for 3k iterations with the combined cross-entropy coverage loss. \newcite{see2017get} use Adagrad \cite{Adagrad} with learning rate 0.15 and an initial accumulator value of 0.1. Early stopping on validation is used to select the final model.

We adopt a simpler procedure, training our full model with the \textit{scratchpad mechanism} for 100k iterations with Adam and a learning rate of $1e^{-4}$ as compared to the two-step procedure in \newcite{see2017get} taking 230k iterations. We follow \newcite{see2017get} in using a batch size of $16$ and clipping gradient norms to 2.

\section{Analysis}
To gain insight into the behaviour and performance of our \textit{Scratchpad Mechanism}, we analyze the output for Question Generation and Translation. We start by conducting a human evaluation study on the Question Generation task, since this task is relatively new and it is well known that quantitative metrics like BLEU do not always correlate with human assessed quality of generated text\footnote{The relation between BLEU scores and more canonical tasks such as machine translation and summarization have already been studied in the literature.\cite{bojar2017findings,graham2015re}}. 
Later we use the attention heatmaps to visualize how the \textit{scratchpad mechanism} drives
the attention weights to be more focused on the relevant source token(s). Additionally, we analize the \textit{entropies} of the attention weights to understand how the \textit{scratchpad mechanism} better allows models to attend to the input. We hypothesize that this is one of the reasons that lead to good performance of
the \textit{scratchpad mechanism} as the decoder ends up being more focused than with 
the standard seq2seq models. 

\subsection{Human Evaluations}
For our human evaluation we use two standard metrics from the machine translation community: \textit{Adequacy} and \textit{Fluency} \cite{bojar2017findings}.
To compute \textit{Adequacy}, human judges are presented with a reference output and the system proposed output, and are asked to rate the adequacy of the proposed output in conveying the meaning of the reference output on a scale from 0-10.
To compute \textit{Fluency}, the judges are asked to rate, on a scale from 0-10, whether the proposed output is a fluent English sentence. We used crowd-sourced judges. Each output is rated by 3 judges.

Table \ref{table:qual_metrics} summarizes the human 
evaluation results for our \textit{Scratchpad Mechanism} and
two more baselines. As the table shows, the judges assigned higher fluency and adequacy scores to our approach than both the coverage-based and copynet baseline. 
In the table we also report the fluency score of the gold questions as a way to measure the gap between the generated questions and the expected ones. Our approach is only 2 points behind the gold when it comes to generation fluency.
\begin{table}[H]
\centering
\begin{tabular}{l c c}
\toprule
Model & Fluency & Adequacy \\ 
\cmidrule(lr){1-1} \cmidrule(lr){2-2} \cmidrule(lr){3-3}

Gold & 9.13 & \xmark \\

\cmidrule(lr){1-3}
Copynet & $5.18$ & $5.23$ \\
+ Coverage & $6.64$ & $6.16$ \\
+ Scratchpad & $\textbf{7.38}$ & $\textbf{6.59}$ \\
\bottomrule
\end{tabular}
\caption{Human evaluations show that the \textit{Scratchpad} delivers a large improvement in both \textit{fluency} and \textit{adequacy} over \textit{Copynet} and \textit{Coverage}, accentuating the improvement in quantitative metrics (Bleu, Rouge, Meteor).}
\label{table:qual_metrics}
\end{table}



\begin{table}[H]
\begin{adjustbox}{width={\columnwidth}}
\begin{tabular}{l r l r}
\toprule
\multicolumn{2}{c}{Scratchpad vs. Copynet} & \multicolumn{2}{c}{Scratchpad vs. Coverage} \\
\cmidrule(lr){1-2} \cmidrule(lr){3-4}
Both Good & $9.26\%$ & Both Good & $15.11\%$ \\
Scratchpad & $37.78\%$ & Scratchpad & $23.80\%$ \\
Copynet & $6.46\%$ & Coverage & $14.99\%$ \\
Both Bad & $46.5\%$ & Both Bad & $43.07\%$ \\
\cmidrule(lr){1-2} \cmidrule(lr){3-4}
Win Rate & $85.39\%$ & Win Rate & $61.36\%$ \\
\bottomrule
\end{tabular}
\caption{The percentage of times judges preferred one result over the other. In a Head-to-Head evaluation the output of \textit{Scratchpad Encoder} is $9$ and $2$ times as likely to be chosen vs. \textit{Copynet} and \textit{Coverage}, respectively. Win rate is the percentage of times Scratchpad was picked
when the judges chose a single winner (not a tie).}
\label{table:head2head}
\end{adjustbox}
\end{table}

Additionally, we design a side-by-side experiment where judges are presented with 2 generated questions from 2 different systems along with the \textit{reference} and asked to judge which output presents a better \textit{paraphrase} to the reference question. Judges take into consideration the grammatical correctness of the question as well as its ability to capture the meaning of the reference question fluently. In Table \ref{table:head2head} We show that in head-to-head evaluations, human judges are nine times as likely to prefer \textit{scratchpad} generated questions over copynet and nearly two times over coverage, accentuating the improved \textit{fluency} and \textit{adequacy} of \textit{scratchpad} generated questions.

\subsection{Attention Visualization and Analysis}
In the standard attention setup, the weights assigned to each encoder 
output is determined by the decoder internal state and the encoder output 
($s^i$ and $h_t$) in Equation 1. Throughout the decoding steps, only $s^i$ varies 
from timestep to the next. Our \textit{scratchpad mechanism} allows the encoder outputs
to change in order to keep track of generated output, so that both $s^i$ and $h_t$ will
vary from timestep to the next, hence more focused attention can be generated.

\begin{figure*}[t]
\captionsetup[subfigure]{labelformat=empty}
\centering
   \begin{subfigure}{}
   \includegraphics[width=\textwidth]{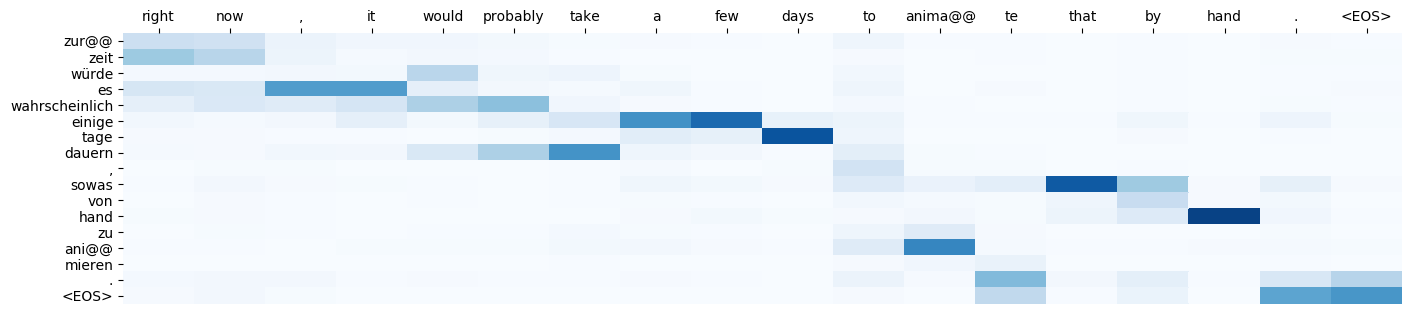}
   \caption{Scratchpad}
    \end{subfigure}
    \begin{subfigure}{}
   \includegraphics[width=\textwidth]{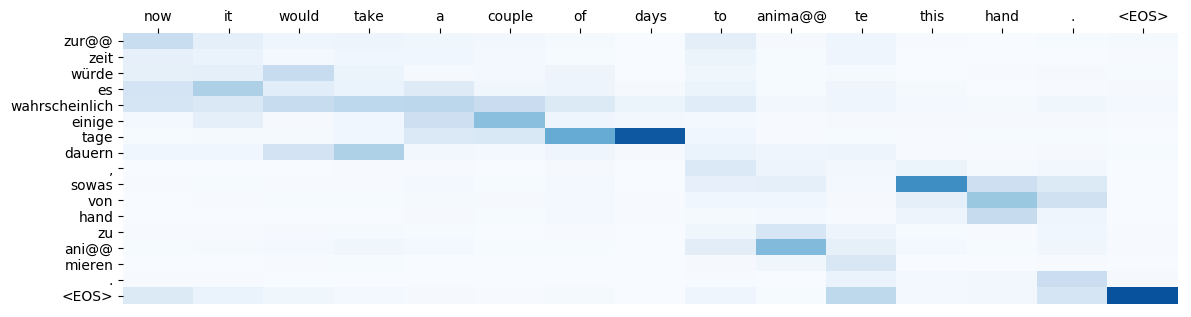}
   \caption{No Scratchpad}
    \end{subfigure}
\caption{Attention over the source sequence visualized at each decoder timestep with and without the scratchpad mechanism. Darker cells mean higher values. "@@" at the end of a bpe token denotes it should be concatenated with the following token(s) to make a word. With Scratchpad, we see sharper attention earlier in the sentence that better aligns with word-level translations (e.g. 'hand' is properly attended to with scratchpad, but not with non-scratchpad). Additionally, some words that are never properly translated (e.g. wahrscheinlich - 'probably') by the non-scratchpad model are not heavily attended to, whereas with the scratchpad mechanism, they are.}
\label{fig:example}
\end{figure*}

We demonstrate this behavior in Fig~\ref{fig:example} where two sentences in 
a German to English machine translation system are shown. In the top figure, 
attention weights are shown when the \textit{scratchpad mechanism} is utilized, while in the bottom 
Figure standard attention is used. As can be seen from the figures, 
attention weights are more focused especially in the first few steps of decoding that better aligns with word-level translations (e.g. 'hand' is properly attended to with scratchpad, but not with non-scratchpad). Additionally, some words that are never properly translated (e.g. wahrscheinlich - 'probably') by the non-scratchpad model are not heavily attended to, whereas with the scratchpad mechanism, they are.

\begin{figure}[tbhp]

\includegraphics[width=\columnwidth]{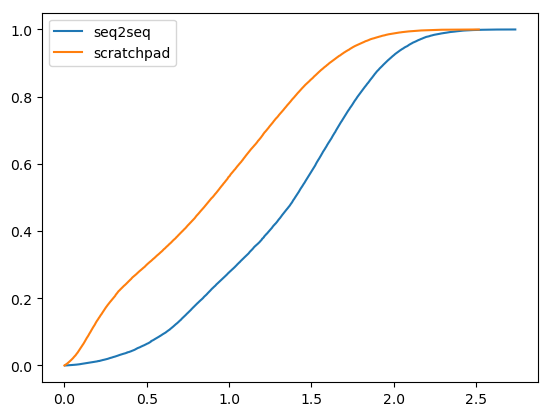}
\caption{We plot the cumulative frequency of attention distribution entropies. On the $Y$-axis is the proportion of attention distribution entropies lower than or equal to $x$.}
\label{fig:cdf}
\end{figure}

We also demonstrate this effect quantitatively. Recall the attention distribution $\at^i_t$ over the input $[x_1,...,x_n]$ generated at each decoding timestep $i$. By calculating the entropy  $\text{ent}^i = -\sum_{t} \at^i_t * \log(\at^i_t)$ and taking the mean of this value across a set of output sentences we can measure how well the model ``focuses'' on input sequences (e.g. $[x_1,...,x_n]$) as it decodes. The \textit{lower} the entropy, the \textit{sharper} the attention distribution. We evaluate this metric on the IWSLT14 De$\rightarrow$En test set for the \textit{scratchpad} and non-scratchpad models. By adding the \textit{scratchpad mechanism}, the mean entropy decreases substantially from $1.33$ to $0.887$ - indicating that it makes the model more selective (focusing on \textit{fewer} input tokens with \textit{higher} weights during generation). Additionally, we plot in  Fig.~\ref{fig:cdf} the cumulative frequency of the word-level entropies $\text{ent}^i$ for all output timesteps $i$. Note from the graph that for every value $x$, the \textit{scratchpad} model produces more attention distributions with an entropy $\leq x$. Finally, the shape of the curve changes to be less sigmoidal, with the proportion of particularly \textit{peaky} or \textit{focused} distributions (very low entropy, e.g. $\leq 0.5$) increasing significantly, over $4\times$ that for the non-scratchpad model. 

Previous work based on coverage based approaches \cite{Coverage,see2017get} either imposed an extra term to the loss function or used an extra vector to keep track of which parts of the input sequences had been attended to, thereby
focusing the attention weights in subsequent steps on tokens that 
received little attention before. In other words, focusing the attention on the relevant parts of the input. Our proposed approach
naturally learns to focus the attention on the important tokens, without a need for modifying the loss function or introducing coverage vectors.

\section{Related work}
\paragraph{Machine Translation}
Since \newcite{Seq2Seq} introduced the sequence-to-sequence
paradigm the approach became 
the defacto standard for performing machine translation.
Improvements over the approach followed, first by the 
introduction of attention \cite{Attention} 
which helped seq2seq translation to focus 
on certain tokens of the encoder outputs. 
Later on, many improvements were described in the Google neural machine translation system \cite{Wu2016GooglesNM}, including utilizing coverage penalty \cite{Coverage} while decoding. The Transformer model was introduced
to alleviate the dependence on RNNs in both the encoder
and the decoder steps
\cite{Vaswani2017AttentionIA}.
Our proposed model sits on top of the seq2seq framework, and could be used with any choice of encoder/decoder as long 
as attention is used.

\paragraph{Summarization}
Since \newcite{Rush2015ANA} first applied neural networks to abstractive text summarization, work has focused on augmenting models \cite{Chopra2016AbstractiveSS, Nallapati2016AbstractiveTS,Copynet}, incorporating syntactic and semantic information \cite{Takase2016NeuralHG}, or direct optimization of the metric at hand \cite{Ranzato2016SequenceLT}. \newcite{Nallapati2016AbstractiveTS} adapted the DeepMind question-answering dataset \cite{Hermann2015TeachingMT} for summarization and provided the first abstractive and extractive \cite{Nallapati2016SummaRuNNerAR} models. \newcite{see2017get} demonstrated that pointer-generator networks can significantly improve the quality of generated summaries. Additionally, work has explored using Reinforcement Learning, often with additional losses or objective functions to improve performance \cite{Hsu2018AUM,Paulus2018ADR,Li2018ActorCriticBT,elikyilmaz2018DeepCA,Pasunuru2018MultiRewardRS}. Finally, \newcite{Gehrmann2018BottomUpAS} demonstrated that a two-stage procedure, where a model first identifies spans in the article that could be copied into the summary which are used to restrict a second pointer-generator model can reap significant gains.
\paragraph{Question Generation}
Early work on translating SPARQL queries into natural language relied heavily on hand-crafted rules \cite{ngonga2013sorry,ngonga2013sparql2nl} or manually crafted templates to map selected categories of SPARQL queries to questions \cite{trivedi2017lc,seyler2017knowledge}.
In \cite{serban2016generating} knowledge base triplets are used to
generate questions using encoder-decoder framework that operates on entity and predicate embeddings trained using TransE \cite{bordes2011learning}. Later, \citet{elsahar2018zero} extended this approach to support unseen predicates. Both approaches operate on triplets, meaning they have limited capability beyond
generating simple questions and cannot generate the far more complex compositional questions that our approach does by operating on the more expressive SPARQL query (logical form).
In the question generation domain, there has been a recent surge in research on generating questions for a given paragraph of text \cite{song2017unified,du2017learning,tang2017question,duan2017question,Wang2018LearningTA,yao2018teaching}, with most of the work being a variant of the seq2seq approach.
In \newcite{song2017unified}, a seq2seq model with copynet
and a coverage mechanism \cite{Coverage} is used to achieve state-of-the-art results. We have demonstrated that our \textit{Scratchpad} outperforms this approach in both quantitative and qualitative evaluations.
\paragraph{Attention}
Closest to our work, in the general paradigm of seq2seq learning, is
the coverage mechanism introduced in \citet{Coverage}
and later adapted  for summarization in \newcite{see2017get}. 
Both works try to minimize erroneous repetitions generated by a copy mechanism by introducing a new vector to keep track of what has been used from the encoder thus far. \citet{Coverage}, for example, use an extra GRU to keep track of this information, whereas \citet{see2017get} keep track of the sum of attention weights and add a penalty to the loss function based on it to discourage
repetition. Our approach is much simpler than either solution since it does not require any extra vectors or an additional loss term; rather, the encoder vector itself is being used as \textit{scratch memory}. Our experiments also show that for the question generation task, the Scratchpad performs better than coverage based approaches. 

Our idea was influenced by the 
dialogue generation work of \citet{eric2017copy} in
which the entire sequence of interactions is re-encoded every time a 
response is generated by the decoder.

\section{Conclusion}
In this paper, we introduce the \textit{Scratchpad Mechanism}, a novel write operator, to the sequence to sequence framework
aimed at addressing many of the common issues encountered by sequence to sequence models and evaluate it on a variety of standard conditional natural language generation tasks. By letting the decoder 'keep notes' on the encoder, or said another way, re-encode the input at every decoding step, the Scratchpad Mechanism  effectively  guides future generation. The Scratchpad Mechanism attains state of the art in Machine Translation, Question Generation, and Summarization on standard metrics and human evaluation across multiple datasets. In addition, our approach decreases training time and model complexity compared to other leading approaches.
Our success on such a diverse set of tasks, input data, and volumes of training data underscores the generalizability of our approach and its conceptual simplicity make it easy to add  to any sequence to sequence model with attention.

\section*{Acknowledgments}
The work was begun while the first author was interning at Apple Siri. We thank Christopher R\'e for his helpful comments and feedback on the paper. We thank the reviewers for insightful comments and suggestions.

\bibliography{acl2019}
\bibliographystyle{acl_natbib}
\end{document}